%% file: MIDORF_PAIN_ARuiz2016.tex
\documentclass[runningheads]{llncs}
\usepackage{graphicx}
\usepackage{amsmath,amssymb} 
\usepackage[table,xcdraw]{xcolor}
\usepackage{color}

\begin{document}
\pagestyle{headings}
\mainmatter

\title{Multi-instance Dynamic Ordinal Random Fields for Weakly-Supervised Pain Intensity Estimation} 
\titlerunning{MI-DORF for weakly-supervised pain intensity estimation}
\authorrunning{Ruiz et al.}

\author{Adria Ruiz$^*$, Ognjen Rudovic$^\dagger$, Xavier Binefa$^*$, Maja Pantic$^{\dagger,\triangledown}$}
\institute{$^*$ DTIC, Universitat Pompeu Fabra, Barcelona \\
           $^\dagger$ Department of Computing, Imperial College London, UK \\
           $^\triangledown$ EEMCS, University of Twente, The Netherlands}

\maketitle

\begin{abstract}
In this paper, we address the Multi-Instance-Learning (MIL) problem when bag labels are naturally represented as ordinal variables (Multi--Instance--Ordinal Regression). Moreover, we consider the case where bags are temporal sequences of ordinal instances. To model this, we propose the novel Multi-Instance Dynamic Ordinal Random Fields (MI-DORF). In this model, we treat instance-labels inside the bag as latent ordinal states. The MIL assumption is modelled by incorporating a high-order cardinality potential relating bag and instance-labels,into the energy function. We show the benefits of the proposed approach on the task of weakly-supervised pain intensity estimation from the UNBC Shoulder-Pain Database. In our experiments, the proposed approach significantly outperforms alternative non-ordinal methods that either ignore the MIL assumption, or do not model dynamic information in target data.

\end{abstract}

\section{Introduction}
\label{sec:introduction}

Mutli-Instance-Learning (MIL) is a popular modelling framework for addressing different weakly-supervised problems \cite{babenko2011robust,wu2014milcut,ruiz2014regularized}. In traditional Single-Instance-Learning (SIL), the fully supervised setting is assumed with the goal to learn a model from a set of feature vectors (instances) each being annotated in terms of target label $y$. By contrast, in MIL, the weak supervision is assumed, thus, the training set is formed by bags (sets of instances), and only labels at bag-level are provided. Furthermore, MIL assumes that there exist an underlying relation between the bag-label (e.g., video) and the labels of its constituent instances (e.g., image frames). In standard Multi-Instance-Classification (MIC) \cite{maron1998framework}, labels are considered binary variables $y \in \{-1,1\}$ and negative bags are assumed to contain only instances with an associated negative label. In contrast, positive bags must contain at least one positive instance. Another MIL assumption is related to the Multi-Instance-Regression (MIR) problem \cite{ray2001multiple}, where $y \in R$ is a real-valued variable and the maximum instance-label within the bag is usually assumed to be equal to $y$. Note, however, that none of these assumptions accounts for structure in the bag labels. Yet, this can be important in case when the bag labels are ordinal, i.e., $y \in \{ 0 \prec ... \prec l \prec L \}$, as in the case of various ratings or intensity estimation tasks. In this work, we focus on the novel modelling task to which we refer as Multi-Instance-Ordinal Regression (MIOR). Similar to MIR, in MIOR we assume that the maximum instance ordinal value within a bag is equal to its label.


To demonstrate the benefits of the proposed approach to MIOR, we apply it to the task of automatic pain estimation \cite{lucey2011painful}. Pain monitoring is particularly important in clinical context, where it can provide an objective measure of the patient's pain level (and, thus, allow for  proper treatment) \cite{Aung2015automatic}. The aim is to predict pain intensity levels from facial expressions (in each frame in a video sequence) of a patient experiencing pain. To obtain the labelled training data, the pain level is usually manually coded on an ordinal scale from low to high intensity \cite{hjermstad2011studies}. To estimate the pain, several SIL methods have been proposed \cite{rudovic2013automatic,kaltwang2012continuous}. Yet, the main limitation of these approaches is they require the frame-based pain level annotations to train the models, which can be very expensive and time-consuming. To reduce the efforts, MIL approaches have recently been proposed for automatic pain detection \cite{wu2015multi,sikka2013weakly,ruiz2014regularized}. Specifically, a weak-label is provided for the whole image sequence (in terms of the maximum observed pain intensity felt by the patient). Then, a video is considered as a bag, and image frames as instances, where the pain labels are provided per bag. In contrast to per-frame annotations, the bag labels are much easier to obtain. For example, using patients self-reports or external observers \cite{lucey2011painful}. Yet, existing MIL approaches for the task focus on the MIC setting, i.e, pain intensities are binarized and model predicts only the presence or absence of pain. Consequently, these approaches are unable to deal with Ordinal Regression problems, and, thus, estimate different intensity levels of pain -- which is critical for real-time pain monitoring. 

In this paper, we propose Multi-Instance Dynamic Ordinal Random Fields (MI-DORF) for MIL with ordinal bag labels. We build our approach using the notion of Hidden Conditional Ordinal Random Fields framework  (HCORF) \cite{kim2010hidden}, for modeling of linear-chains of ordinal latent variables. In contrast to HCORF that follows the Single-Instance paradigm, the energy function employed in MI-DORF is designed to model the MIOR assumption relating instance and bag labels. In relation to static MIL methods, our MI-DORF also incorporates dynamics within the instances, encoded by transitions between ordinal latent states. This information is useful when instances (frames) in a bag are temporally correlated, as in pain videos. The main contributions of this work can be  summarised as follows:

\begin{itemize}
\renewcommand\labelitemi{$\bullet$}

\item To the best our knowledge, the proposed MI-DORF is the first MIL approach that imposes ordinal structure on the bag labels. The proposed method also incorporates dynamic information that is important when modeling temporal structure in instances within the bags (i.e., image sequences). While modeling the temporal structure has been attempted in \cite{wu2015multi,liu2015video}, there are virtually no works that account for both ordinal and temporal data structures within MIL framework. 

\item We introduce an efficient inference method in our MI-DORF, which has a similar computational complexity as the forward-backward algorithm \cite{barber2012bayesian} used in standard first-order Latent-Dynamic Models (e.g HCORF). This is despite the fact that we model high-order potentials modelling the Multi-Instance assumption.

\item We show in the task of automated pain intensity estimation from the UNBC Shoulder-Pain Database \cite{lucey2011painful} that the proposed MI-DORF outperforms significantly existing related approaches applicable to this task. We show that due to the modeling of the ordinal and temporal structure in the target data, we can infer instance-level pain intensity levels that largely correlate with manually obtained frame-based pain levels. Note that we do so by using only the bag labels for learning, that are easy to obtain. To our knowledge, this has not been attempted before.
\end{itemize}

\section{Related Work}
\label{sec:related}

\textbf{Multi-Instance-Learning.} Existing  MIC/MIR approaches usually follow the bag-based or instance-based paradigms \cite{amores2013multiple}. In bag-based methods, a feature vector representation for each bag is first extracted. Then, these representations are used to train standard Single-Instance Classification or Regression methods, used to estimate the bag labels. Examples include Multi-Instance Kernel \cite{gartner2002multi}, MILES \cite{chen2006miles}, MI-Graph \cite{zhou2009multi} and MI-Cluster Regression \cite{wagstaff2008multiple}. The main limitation of these approaches is that the learned models can only make predictions at the bag-level. However, these methods cannot work in in the weakly-supervised settings, where the goal is to predict instance-labels (e.g., frame-level pain intensity) from a bag (e.g., a video). In contrast, instance-based methods directly learn classifiers which operate at the instance level. For this, MIL assumptions are incorporated into the model by considering instance-labels as latent variables. Examples include Multi-Instance Support Vector Machines \cite{andrews2002support} (MI-SVM), MILBoost \cite{zhang2005multiple}, and Multi-Instance Logistic Regression \cite{hsu2014augmented}. The proposed MI-DORF model follows the instance-based paradigm by treating instance-labels as ordinal latent states in a Latent-Dynamic Model. In particular, it follows a similar idea to that in the Multi-Instance Discriminative Markov Networks \cite{hajimirsadeghi2013multiple}. In this approach, the energy function of a Markov Network is defined by using cardinality potentials modelling the relation between bag and instance labels. MI-DORF also make use of cardinality potentials, however, in contrast to the works described above, it accounts for the ordinal structure at both the bag and instance level, while also accounting for the dynamics in the latter.

\textbf{Latent-Dynamic Models.} Popular methods for sequence classification are Latent-Dynamic Models such as Hidden Conditional Random Fields (HCRFs) \cite{quattoni2007hidden} or Hidden-Markov-Models (HMMs) \cite{rabiner1986introduction}. These methods are variants of Dynamic Bayesian Networks (DBNs) where a set of latent states are used to model the conditional distribution of observations given the sequence label. In these approaches, dynamic information is modelled by incorporating probabilistic dependence between time-consecutive latent states. MI-DORF builds upon the HCORF framework \cite{kim2010hidden} which considers latent states as ordinal variables. However, HMM and HCRF/HCORF follow the SIL paradigm where the main goal is to predict sequence labels. In contrast, in MI-DORF, we define a novel energy function that encodes the MI relationship between the bag labels, and also their latent ordinal states. Note also that the recent works (e.g., \cite{wu2015multi}, \cite{liu2015video}) extended HMMs/HCRFs, respectively, for MIC. The reported results in this work suggested that modeling dynamics in MIL can be beneficial when bag-instances exhibit temporal structure. However, these methods limit their consideration to the case where bag labels are binary and, therefore, are unable to solve the MIOR problem.

\textbf{MIL for weakly-supervised pain detection.} Several works attempted pain detection in the context of the weakly-supervised MIL. As explained in Sec.\ref{sec:introduction}, these approaches adopt the MIC framework where pain intensities are binarized. For instance, \cite{sikka2013weakly} proposed to extract a Bag-of-Words representation from video segments and treat them as bag-instances. Then, MILBoosting \cite{zhang2005multiple} was applied to predict sequence-labels under the MIC assumption. Following the bag-based paradigm, \cite{ruiz2014regularized} developed the Regularized Multi-Concept MIL method capable of discovering different discriminative pain expressions within an image sequence. More recently, \cite{wu2015multi} proposed  MI Hidden Markov Models, an adaptation of standard HMM to the MIL problem. The limitation of these approaches is that they focus on the binary detection problem, and, thus, are unable to deal with (ordinal) multi-class problems (i.e., pain intensity estimation). This is successfully attained by the proposed MI-DORF.

\section{Multi-Instance Dynamic Ordinal Random Fields (MI-DORF)}
\label{sec:model}

\subsection{Multi Instance Ordinal Regression (MIOR)}
\label{sec:problem_description}
In the MIOR weakly-supervised setting, we are provided with a training set $\mathcal{T}=\{(\mathbf{X}_1,y_1),(\mathbf{X}_2,y_2),...,(\mathbf{X}_N,y_N)\}$ formed by pairs of structured-inputs  $X\in\mathcal{X}$ and labels $y \in \{ 0 \prec ... \prec l \prec L \}$ belonging to a set of $L$ possible ordinal values.  In this work, we focus on the case where  $\mathbf{X} = \{ \mathbf{x}_{1},\mathbf{x}_{2},...,\mathbf{x}_{T} \}$ are temporal sequences of $T$ observations $\mathbf{x} \in R^d$ in a d-dimensional space   \footnote{Total number of observations $T$ can vary across different sequences}. Given the training-set $\mathcal{T}$, the goal is to learn a model $\mathcal{F}: \mathcal{X} \rightarrow \mathcal{H}$ mapping sequences $\mathbf{X}$ to an structured-output $\mathbf{h} \in \mathcal{H}$. Concretely, $\mathbf{h} = \{ h_{1},h_{2},...,h_{T} \}$ is a sequence of variables $h_t \in \{ 0 \prec ... \prec l \prec L \}$ assigning one ordinal value for each observation $\mathbf{x_t}$ . In order to learn the model $\mathcal{F}$ from $\mathcal{T}$, MIOR assumes that the maximum ordinal value in $\mathbf{h}_n$ must be equal to the label $y_n$ for all sequences $\mathbf{X}_n$:

\begin{equation}
\label{eq:miorassumption}
\mathcal{F}( \mathbf{X}_n ) = \mathbf{h}_n \hspace{3mm} s.t \hspace{3mm} y_n = \max_h(\mathbf{h}_n) \hspace{6mm} \forall \hspace{1mm} (\mathbf{X}_n,y_n) \in  \mathcal{T}
\end{equation}

\begin{figure}[t]
    \centering
        \includegraphics[width=1\linewidth]{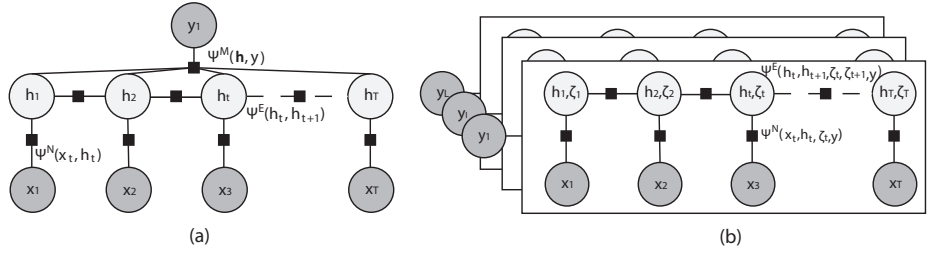}
    \caption{(a) Graphical representation of the proposed MI-DORF model. Node potentials $\Psi^{N}$ model the compatibility between a given observation $\mathbf{x}_t$ and a latent ordinal value $h_t$ . Edge potentials $\Psi^E$ take into account the transition between consecutive latent ordinal states $h_t$ and $h_{t+1}$. Finally, the high-order cardinality potential $\Psi^{M}$ models the MIOR assumption relating all the latent ordinal states $\mathbf{h}_t$  with the bag-label $y$. (b) Equivalent model defined using the auxiliary variables $\zeta_t$ for each latent ordinal state. The use of these auxiliary variables and the redefinition of node and edge potentials allows us to perform efficient inference over the MI-DORF model (see Sec. \ref{sec:inference}).}
    \label{fig:mildorf}
     \vspace{-0.3cm}
\end{figure}

\subsection{MI-DORF: Model Overview}

 We model the structured-output $\mathbf{h} \in \mathcal{H}$ as a set of ordinal latent variables. We then define the conditional distribution of $y$ given observations $\mathbf{X}$. Formally, $P(y|\mathbf{X};\theta)$ is assumed to follow a Gibbs distribution as:

\begin{equation}
\label{eq:cond_probability}
P(y|\mathbf{X};\theta) = \frac{\sum_h{e^{-\Psi(\mathbf{X},\mathbf{h},y;\theta)}}}{\sum_{y^\prime}\sum_h{e^{-\Psi(\mathbf{X},\mathbf{h},y^\prime;\theta)}}},
\end{equation}
where $\theta$ is the set of the model parameters. As defined in Eq. \ref{eq:energy_func}, the energy function $\Psi$ defining the Gibbs distribution is composed of the sum of three different types of potentials. An overview of the model is shown in Fig. \ref{fig:mildorf}(a).

\begin{equation}
\label{eq:energy_func}
 \Psi(\mathbf{X},\mathbf{h},y;\theta) = \sum_{t=1}^{T} \Psi^{N}(\mathbf{x}_t,h_t;\theta^N) + \sum_{t=1}^{T-1} \Psi^{E}(h_t,h_{t+1};\theta^E) + \Psi^{M}(\mathbf{h},y,\theta^{M}),
\end{equation}

\subsubsection{MI-DORF: Ordinal node potentials} The node potentials $\Psi^{N}(\mathbf{x},h;\theta^N)$ aim to capture the compatibility between a given observation $\mathbf{x}_t$ and the latent ordinal value $h_t$. Similar to HCORF, it is defined using the ordinal likelihood model \cite{winkelmann2006analysis}:

\begin{equation}
  \label{eq:ordinal_regression}
  \Psi^{N}(\mathbf{x},h=l;\theta^N)= \log \Bigg(  \Phi \bigg( \frac{b_l - \mathbf{\beta}^T \mathbf{x})}{\sigma}\bigg) - \Phi \bigg( \frac{b_{(l-1)} - \mathbf{\beta}^T \mathbf{x})}{\sigma}\bigg) \Bigg),
\end{equation}
where $\Phi(\cdot)$ is the normal cumulative distribution function (CDF), and $\theta^N=\{\beta,\mathbf{b},\sigma\}$ is the set of potential parameters. Specifically, the vector $\beta \in {R}^d$ projects observations $\mathbf{x}$ onto an ordinal line divided by a set of cut-off points ${b_0} =  - \infty  \le  \cdots  \le {b_L} = \infty $. Every pair of contiguous cut-off points divide the projection values into different bins corresponding to the different ordinal states $l=1,...,L$. The difference between the two CDFs provides the probability of the latent state $l$ given the observation $\mathbf{x}$, where $\sigma$ is the standard deviation of a Gaussian noise contaminating the ideal model (see \cite{kim2010hidden} for details). In our case, we  fix $\sigma=1$, to avoid model over-parametrization.

\subsubsection{MI-DORF: Edge potentials} The edge potential $\Psi^E(h_t,h_{t+1};\theta^E)$ models temporal information regarding compatibilities between consecutive latent ordinal states as:

\begin{equation}
 \Psi^E(h_t=l,h_{t+1}=l^\prime;\theta^E) = \mathbf{W}_{l,l^\prime},
\end{equation}
where $\theta^E={\mathbf{W}^{L \times L}}$ represents a real-valued transition matrix, as in standard HCORF. The main goal of this potential is to perform temporal smoothing of the instance intensity levels.

\subsubsection{MI-DORF: Multi-Instance-Ordinal potential}  In order to model the MIOR assumption (see Eq. \ref{eq:miorassumption}), we define a high-order potential $\Psi^{M}(\mathbf{h},y;\theta^M)$ involving label $y$ and all the sequence latent variables $\mathbf{h}$ as:

\begin{equation}
\label{eq:mil_potential}
 \Psi^{M}(\mathbf{h},y;\theta^M) = 
 \begin{cases}
            w \sum_{t=1}^{T} \mathbf{I} (h_t==y) \hspace{3mm} \text{iff} \hspace{3mm} \max (\mathbf{h}) = y \\
            -\infty \hspace{3mm} \text{otherwise}
    \end{cases},
\end{equation}
where $\mathbf{I}$ is the indicator function, and $\theta^M=w$. Note that when the maximum value within $\mathbf{h}$ is not equal to $y$, the energy function is equal to $-\infty$ and, thus, the probability $P(y|\mathbf{X};\theta)$ drops to 0. On the other hand, if the MI assumption is fulfilled, the summation $w \sum_{t=1}^{T} \mathbf{I} (h_t==y)$ increases the energy proportionally to $w$ and the number of  latent states $\mathbf{h} \in h_t$ that are equal to $y$. This is convenient since, in sequences annotated with a particular label, it is more likely to find many latent ordinal states with such ordinal level. Therefore, the defined MI potential does not only model the MI-OR assumption but also provides mechanisms to learn how important is the proportion of latent states $\mathbf{h}$ that are equal to the label. Eq. \ref{eq:mil_potential} is a special case of  cardinality potentials \cite{gupta2007efficient} also employed in binary Multi-Instance Classification \cite{hajimirsadeghi2013multiple}.

\subsection{MI-DORF: Learning}
\label{sec:training}

Given a training set $\mathcal{T}=\{(\mathbf{X}_1,y_1),(\mathbf{X}_2,y_2),...,(\mathbf{X}_N,y_N)\}$, we learn the model parameters $\theta$ by minimizing the regularized log-likelihood:
\begin{equation}
\label{eq:training}
 \min_\mathbf{\theta} \hspace{3mm} \sum_{i=1}^N \log P(y|\mathbf{X};\theta) + \mathcal{R}(\theta),
\end{equation}
where the regularization function $\mathcal{R}(\theta)$ over the model parameters is defined as:
\begin{equation}
 \mathcal{R}(\theta) = \alpha (||\beta||_2^2 + ||\mathbf{W}||_F^2)
\end{equation}
and $\alpha$ is set via a validation procedure. The objective function in Eq.\ref{eq:training} is differentiable and standard gradient descent methods can be applied for optimization. To this end, we use the L-BFGS Quasi-Newton method \cite{byrd1994representations}. The gradient evaluation involves marginal probabilities $p(h_t|\mathbf{X})$ and $p(h_t,h_{t+1}|\mathbf{X})$ which can be efficiently computed using the proposed algorithm in Sec. \ref{sec:inference}.

\subsection{MI-DORF: Inference}
\label{sec:inference}

The evaluation of the conditional probability $P(y|\mathbf{X};\theta)$ in Eq.\ref{eq:cond_probability} requires computing $\sum_h{e^{-\Psi(\mathbf{X},\mathbf{h},y;\theta)}}$ for each label $y$. Given the exponential number of possible latent states $\mathbf{h} \in \mathcal{H}$, efficient inference algorithms need to be used. In the case of  Latent-Dynamic Models such as HCRF/HCORF, the forward-backward algorithm \cite{barber2012bayesian} can be applied. This is because the pair-wise linear-chain connectivity between latent states $\mathbf{h}$. However, in the case of MI-DORF, the inclusion of the cardinality potential $\Psi^{M}(\mathbf{h},y;\theta^{M})$ introduces a high-order dependence between the label $y$ and all the latent states in $\mathbf{h}$. Inference methods with cardinality potentials has been previously proposed in \cite{gupta2007efficient,tarlow2012fast}. However, these algorithms only consider the case where latent variables are independent and, therefore, they can not be applied in MI-DORF. For these reasons, we propose an specific inference method. The idea behind it is to apply the standard forward-backward algorithm by converting the energy function defined in Eq. \ref{eq:energy_func} into an equivalent one preserving the linear-chain connectivity between latent states $\mathbf{h}$.  

To this end, we introduce a new set of auxiliary variables $\boldsymbol{\zeta} = \{\zeta_1,\zeta_2,...,\zeta_T\}$, where each $\zeta_t \in \{0,1\}$ takes a binary value denoting whether the sub-sequence $\mathbf{h}_{1:t}$ contains at least one ordinal state $h$ equal to $y$. Now we redefine the MI-DORF energy function in Eq. \ref{eq:energy_func} as:
\begin{equation}
\label{eq:energy_func_inference}
 \Psi(\mathbf{X},\mathbf{h},\boldsymbol{\zeta},y;\theta) = \sum_{t=1}^{T} \Psi^{N}(\mathbf{x}_t,h_t,\zeta_t,y;\theta^N) + \sum_{t=1}^{T-1} \Psi^{E}(h_t,h_{t+1},\zeta_t,\zeta_{t+1},y;\theta^E),
\end{equation}
where the new node and edge potentials are given by:
 \begin{equation}
  \label{eq:inference_node_potential}
  \Psi^{N}(\mathbf{x}_t,h_t,\zeta_t,y;\theta^N) = 
   \begin{cases}
            \Psi^{N}(\mathbf{x}_t,h_t;\theta^N) + w\mathbf{I} (h_t==y) \hspace{2mm} \text{iff} \hspace{2mm} h_t <= y \\
            -\infty \hspace{3mm} \text{otherwise}
    \end{cases},
\end{equation}
\begin{equation}
\Psi^{E}(h_t,h_{t+1},\zeta_t,\zeta_{t+1},y;\theta^E) = 
 \begin{cases}
            \mathbf{W}_{h_t,h_{(t+1)}} \hspace{3mm} \text{iff} \hspace{3mm} \zeta_t=0 \land \zeta_{t+1}=0 \land h_{t+1} \neq y\\
            \mathbf{W}_{h_t,h_{(t+1)}} \hspace{3mm} \text{iff} \hspace{3mm} \zeta_t=0 \land \zeta_{t+1}=1 \land h_{t+1} = y \\
            \mathbf{W}_{h_t,h^{(t+1)}} \hspace{3mm} \text{iff} \hspace{3mm} \zeta_t=1 \land \zeta_{t+1}=1 \\
            -\infty \hspace{3mm} \text{otherwise}
  \end{cases}
\end{equation}

 Note that Eq. \ref{eq:energy_func_inference} does not include the MIO potential and, thus, the high-order dependence between the label $y$ and latent ordinal-states $\mathbf{h}$ is removed.  The graphical representation of MI-DORF with the redefined energy function is illustrated in Fig.\ref{fig:mildorf}(b). In order to show the equivalence between energies in Eqs. \ref{eq:energy_func} and \ref{eq:energy_func_inference}, we explain how the the original Multi-Instance-Ordinal potential $\Psi^M$ is incorporated into the new edge and temporal potentials. Firstly, note that $\Psi^{N}$ now also takes into account the proportion of ordinal variables $h_t$ that are equal to the sequence label. Moreover, it enforces $\mathbf{h}$ not to contain any $h_t$ greater than $y$, thus aligning the bag and (max) instance labels. However, the original Multi-Instance-Ordinal potential also constrained $\mathbf{h}$ to contain at least one $h_t$ with the same ordinal value than $y$. This is achieved by using the set of auxiliary variables $\zeta_t$ and the re-defined edge potential $\Psi^{E}$. In this case, transitions between latent ordinal states are modelled but also between auxiliary variables $\zeta_t$. Specifically, when the ordinal state in $h_{t+1}$ is equal to $y$, the sub-sequence $\mathbf{h}_{1:t+1}$ fulfills the MIOR assumption and, thus, $\zeta_{t+1}$ is forced to be $1$. By defining the special cases at the beginning and the end of the sequence ($t=1$ and $t=T$):

 \begin{equation}
  \label{eq:inference_node_potential0}
  \Psi^{N}(\mathbf{x}_1,h_1,,\zeta_1,y) = 
   \begin{cases}
            \Psi^{N}(\mathbf{x}_1,h_1) + w\mathbf{I} (h_1==y) \hspace{2mm} \text{iff} \hspace{2mm} \zeta_1 = 0 \land l_1 < y \\
            \Psi^{N}(\mathbf{x}_1,h_1) + w\mathbf{I} (h_1==y) \hspace{2mm} \text{iff} \hspace{2mm} \zeta_1 = 1 \land l_1 = y \\
            -\infty \hspace{3mm} \text{otherwise}
    \end{cases},
\end{equation}

 \begin{equation}
  \label{eq:inference_node_potentialT}
  \Psi^{N}(\mathbf{x}_T,h_T,\zeta_T,y) = 
   \begin{cases}
            \Psi^{N}(\mathbf{x}_T,h_T) + w\mathbf{I} (h_T==y) \hspace{1.5mm} \text{iff} \hspace{1.5mm} \zeta_T = 1 \land h_T <= y  \\
            -\infty \hspace{3mm} \text{otherwise}
    \end{cases},
\end{equation}

we can see that the energy is $-\infty$ when the MIOR assumption is not fulfilled. Otherwise, it has the same value than the  one defined in Eq.\ref{eq:energy_func} since no additional information is given. The advantage of using this equivalent energy function is that the standard forward-backward algorithm can be applied to efficiently compute the conditional probability:
\begin{equation}
\label{eq:cond_probability2}
P(y|\mathbf{X};\theta) = \frac{\sum_\mathbf{h} \sum_{\boldsymbol{\zeta}} {e^{-\Psi(\mathbf{X},\mathbf{h},\boldsymbol{\zeta},y;\theta)}}}{\sum_{y^\prime}\sum_\mathbf{h} \sum_{\boldsymbol{\zeta}} {e^{-\Psi(\mathbf{X},\mathbf{h},\boldsymbol{\zeta},y^\prime;\theta)}}},
\end{equation}

The proposed procedure has a computational complexity of $\mathcal{O}(T \cdot (2L)^2)$ compared with $\mathcal{O}(T \cdot L^2)$ using standard forward-backward in traditional linear-chain latent dynamical models. Since typically $L<<T$, this can be considered a similar complexity in practice. The presented algorithm can also be applied to compute the marginal probabilities $p(h_t|\mathbf{X})$ and $p(h_t,h_{t+1}|\mathbf{X})$. This probabilities are used during training for gradient evaluation and during testing to predict ordinal labels at the instance and bag level.


\section{Experiments}

\subsection{Baselines and evaluation metrics}
\label{sec:baselines}
The introduced MI-DORF approach is designed to address the Multi-Instance-Ordinal Regression when bags are structured as temporal sequences of ordinal states. Given that this has not been attempted before, we compare MI-DORF with different approaches that either ignore the MIL assumption (Single-Instance) or do not model dynamic information (Static):

\textbf{Single-Instance Ordinal Regression (SIL-OR):} MIL can be posed as a SIL problem with noisy labels. The main assumption is that the majority of instances will have the same label than their bag. In order to test this assumption, we train standard Ordinal Regression \cite{winkelmann2006analysis} at instance-level by setting all their labels to the same value as their corresponding bag. During testing, bag-label is set to the maximum value predicted for all its instances. Note that this baseline can be considered an Static-SIL approach.

\textbf{Static Multi-Instance Ordinal Regression (MI-OR):} Given that no MIOR methods have previously been proposed for this task, we implemented this static approach following the MIOR assumption. This method is inspired by MI-SVM \cite{andrews2002support}, where instance labels are considered latent variables and are iteratively optimized during training. To initialize the parameters of the ordinal regressor, we follow the same procedure as described above in SIL-OR. Then, ordinal values for each instance are predicted and modified so that the MIOR assumption is fulfilled for each bag. Note that if all the predictions within a bag are lower than its label, the instances with the maximum value are set to the bag-label. On the other hand, all the predictions greater than the bag-label are decreased to this value. With this modified labels, Ordinal Regression is applied again and this procedure is applied iteratively until convergence.

\textbf{Multi-Instance-Regression (MIR):} Several methods have been proposed in the literature to solve the MIL problem when bags are real-valued variables. In order to evaluate the performance of this approach in MIOR, we have implemented a similar method as used in \cite{hsu2014augmented}. Specifically, a linear regressor at the instance-level is trained by optimizing a loss function over the bag-labels. This loss models the MIR assumption by using a soft-max function which approximates the maximum instance label within a bag predicted by the linear regressor. Note that a similar approach is also applied in Multi-Instance Logistic Regression \cite{ray2005supervised}. In these works, a logistic loss is used because instance labels take values between 0 and 1. However, we use a squared-error loss to take into account the different ordinal levels.

\textbf{Multi-Instance HCRF (MI-HCRF):} This approach is similar to the proposed MI-DORF. However, MI-HCRF ignores the ordinal nature of labels and models them as nominal variables. For this purpose, we replace the MI-DORF node potentials by a multinomial logistic regression model \footnote{The potential with the Multinomial Logistic Regession model is defined as $\log ( \frac{ \exp(\beta^T_l x)}{ \sum_{ l^\prime \in L} \exp(\beta^T_{l^\prime} x) }   )$ . Where all $\mathbf{\beta_l}$ defines a linear projection for each possible ordinal value $l$ \cite{walecki2015variablestate} }. Inference in MI-HCRF is performed by using the algorithm described in Sec. \ref{sec:inference}.

\textbf{Single-Instance Latent-Dynamic Models (HCRF/HCORF):} We also evaluate the performance of HCRF and HCORF. For this purpose, the Mutli-Instance-Ordinal potential in MI-DORF is replaced by the employed in standard HCRF \cite{quattoni2007hidden}. This potential models the compatibility of hidden state values $\mathbf{h}$ with the sequence-label $y$ but ignores the Multi-Instance assumption. For HCRF, we also replace the node potential as in the case of MI-HCRF. Inference is performed using the standard forward-backward algorithm.

\textbf{Evaluation metrics:} In order to evaluate the performance of MI-DORF and the compared methods, we report results in terms of instance and bag-labels predictions. Note that in the MIL literature, results are usually reported only at bag-level. However, in problems such as weakly-supervised pain detection, the main goal is to predict instance labels (frame-level pain intensities). Given the ordinal nature of the labels, the reported metrics are the Pearson's Correlation (CORR), Intra-Class-Correlation (ICC) and Mean-Average-Error (MAE). For bag-label predictions, we also report the Accuracy and average F1-score as discrete metrics. 

\begin{figure}[t]
  \begin{center}
  \includegraphics[width=1\linewidth]{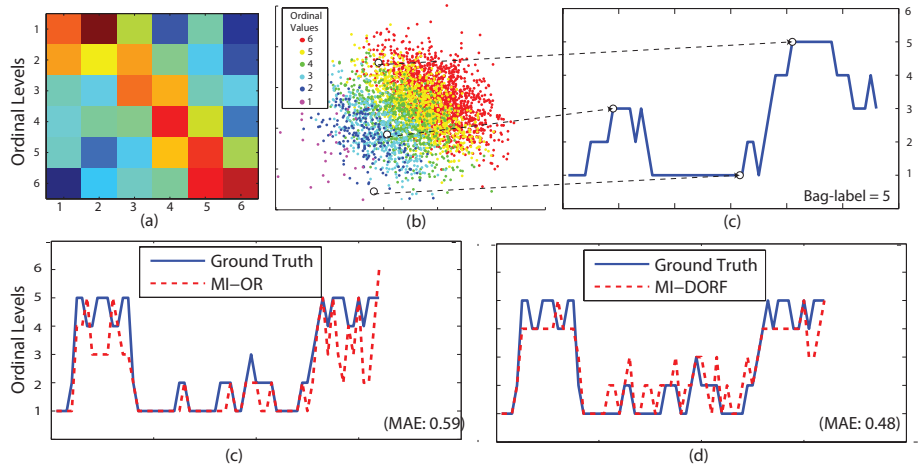}
  \end{center}
  \caption{Description of the procedure used to generate synthetic sequences. (a) A random matrix modelling transition probabilities between consecutive latent ordinal values. (b) Ordinal levels assigned to the random feature vectors according to the ordinal regressor. (c) Example of a sequence of ordinal values obtained using the generated transition matrix. The feature vector representing each observation is randomly chosen between the samples in (b) according to the probability for each ordinal level. (c-d) Examples of instance-level predictions in a sequence for MI-OR and MI-DORF. }
  \label{fig:unbc_results_synthetic}
  \vspace{-0.3cm}
\end{figure}

\input{baselineresults}

\subsection{Synthetic Experiments}
\label{sec:synth_experiments}
\textbf{Synthetic Data:} Given that no standard benchmarks are available for MIOR, we have generated synthetic data. To create the synthetic sequences, we firstly generated a sequence of ordinal values using a random transition matrix. It represents the transition probabilities between temporally-consecutive ordinal levels. The first value for the sequence  is randomly chosen with equal probability among all possible ordinal levels. Secondly, we generate random parameters of an Ordinal Regressor as defined in Eq. \ref{eq:ordinal_regression}. This regressor is used to compute the probabilities for each ordinal level in a set o feature-vectors randomly sampled from a Gaussian distribution. Thirdly, the corresponding sequence observation for each latent state in the sequence is randomly chosen between the sampled feature vectors according to the obtained probability for each ordinal value. Finally, the sequence-label is set to the maximum ordinal state within the sequence following the MIOR assummption, and Gaussian noise ($\sigma=0.25$) is added to the feature vectors. Fig. \ref{fig:unbc_results_synthetic}(a-c) illustrates this procedure. Following this strategy, we have generated  ten different data sets by varying the ordinal regressor parameters and transition matrix. Concretely, each dataset is composed of 100 sequences for training, 150 for testing and 50 for validation. The last set is used to optimize the regularization parameters of each method. The sequences have a variable length between 50 and 75 instances. The dimensionality of the feature vectors was set to 10 and the number of ordinal values to 6. 

\textbf{Results and Discussion:} Table \ref{tab:synth_results} shows the results computed as the average performance over the ten datasets. SIL methods (SIL-OR, HCRF and HCORF ) obtain worse performance than their corresponding MI versions (MI-OR,MI-HCRF and MI-DORF) in most of the evaluated metrics. This is expected since SIL approaches ignore the Multi-Instance assumption. Moreover, HCORF and MI-DORF obtain better performance compared to HCRF and MI-HCRF. This is because the former model the latent states as nominal variables, thus, ignoring their ordinal nature. Finally, note that MI-DORF outperforms the static methods MI-OR and MIR. Although these approaches use the Multi-Instance assumption and incorporate the labels ordering, they do not take into account temporal information. In contrast, MI-DORF is able to model the dynamics of latent ordinal states and use this information to make better predictions when sequence observations are noisy. As Fig. \ref{fig:unbc_results_synthetic}(c-d) shows, MI-OR predictions tends to be less smooth because dynamic information is not taken into account. In contrast, MI-DORF better estimate the actual ordinal levels by modelling transition probabilities between consecutive ordinal levels.

\subsection{Weakly-supervised pain intensity estimation}
\label{sec:unbc_experiments}
In this experiment, we test the performance of the proposed model for weakly-supervised pain intensity estimation. To this end, we use the UNBC Shoulder-Pain Database \cite{lucey2011painful}. This dataset contains recordings of different subjects performing active and passive arm movements during rehabilitation sessions. Each video is annotated according to the maximum pain felt by the patient during the recording in an ordinal scale between 0 (no pain) and 5 (strong pain). These annotations are used as the bag label in the MIOR task. Moreover, pain intensities are also annotated at frame-level in terms of the PSPI scale \cite{prkachin1992consistency}. This ordinal scale ranges from 0 to 15. Frame PSPI annotations are  normalized between 0 and 5, in order to align the scale with the one provided at the sequence level. Furthermore, we used a total of 157 sequences from 25 subjects. The remaining 43 were removed because a high discrepancy between sequence and frame-level annotations was observed. Concretely, we do not consider the cases where the sequence label is 0 and frame annotations contains higher pain levels. Similarly, we also remove sequences with a high-discrepancy in the opposite way. Given the different scales used in frame and sequence annotations, we computed the agreement between them. For this purpose, we firstly obtained the maximum pain intensities at frame-level for all the used sequences. Then, we computed the CORR and ICC between them and their corresponding sequence labels. The results were 0.83 for CORR, and 0.78 in the case of ICC. This high agreement indicates that predictions in both scales are comparable. More importantly, this supports our hypothesis that  sequence labels are highly correlated with frame labels; thus, the used bag labels provide sufficient information for learning the instance labels in our weakly-supervised setting.

\textbf{Facial-features:} For each video frame, we compute a geometry-based facial-descriptor as follows.
Firstly, we obtain a set of 49 landmark facial-points with the method described in \cite{XiongD13}. Then, the obtained points locations are aligned with a mean-shape using Procrustes Analysis. Finally, we generate the facial descriptor by concatenating the $x$ and $y$ coordinates of all the aligned points. According to the MIL terminology, these facial-descriptors are considered the instances in the bag (video).

\textbf{Experimental setup:} We perform Leave-One-Subject-Out Cross Validation similar to \cite{sikka2013weakly}. In each cycle, we use 20 subjects for training, 1 for testing and 4 for validation. This last subset is used to cross-validate the regularization parameters of each particular method. In order to reduce computational complexity and redundant information between temporal consecutive frames, we have segmented the sequences using non-overlapping windows of 0.5 seconds, similar to \cite{sikka2013weakly}. The instance representing each segment is computed as the mean of its corresponding facial-descriptors. Apart from the baselines described in Sec. \ref{sec:baselines}, we also evaluate the performance of the MIC approach considering pain levels as binary variables. For this purpose, we have implemented the MILBoosting \cite{zhang2005multiple} method used in \cite{sikka2013weakly} and considered videos with a pain label greater than 0 as positive. Given that MI-Classification methods are only able to make binary predictions, we use the output probability as indicator of intensity levels, at bag and instance-level, i.e., the output probability is normalized between 0 and 5.

\input{unbcresults}

\textbf{Results and discussion:}  Table \ref{tab:unbc_results} shows the results obtained by the evaluated methods following the experimental setup previously described. By looking into the results of the compared methods, we can derive the following conclusions. Firstly, SI approaches ( SIL-OR, HCORF and HCRF) obtain worse performance than MI-OR and MIR. This is because pain events are typically very sparse in these sequences and most frames have intensity level 0 (neutral). Therefore, the use of the MIL assumption has a critical importance in this problem. Secondly, poor results are obtained by HCRF and MI-HCRF. This can be explained because these approaches consider pain levels as nominal variables and are ignorant of the ordering information of the different pain intensities. Finally, MILBoost trained with binary labels also obtains low performance compared to the MI-OR and MIR. This suggest that current approaches posing weakly-supervised pain detection as a MIC are suboptimal, thus, unable to predict accurately the target pain intensities. By contrast, MI-DORF obtains the best performance across all the evaluated metrics at both the sequence and frame-level. We attribute this to the fact the MI-DORF models the MIL assumption with ordinal variables. Moreover, the improvement of MI-DORF compared to the static approaches, such as MI-OR and MIR, suggests that modelling dynamic information is beneficial in this task. To get better insights into the performance of our weakly supervised approach, we compare its results (in terms of ICC) to those obtained by the fully supervised (at the frame level) state-of-the-art approach to pain intensity estimation - Context-sensitive Dynamic Ordinal Regression \cite{rudovic2015context}. While this approach achieves an ICC of 0.67/0.59, using context/no-context features, respectively, our MI-DORF achieves an ICC of 0.40 without ever seeing the frame labels. This is a good trade-off between the need for the "very-expensive-to-obtain" frame-level annotation, and the model's performance. 

\begin{figure}[t]
  \begin{center}
  \includegraphics[width=1\linewidth]{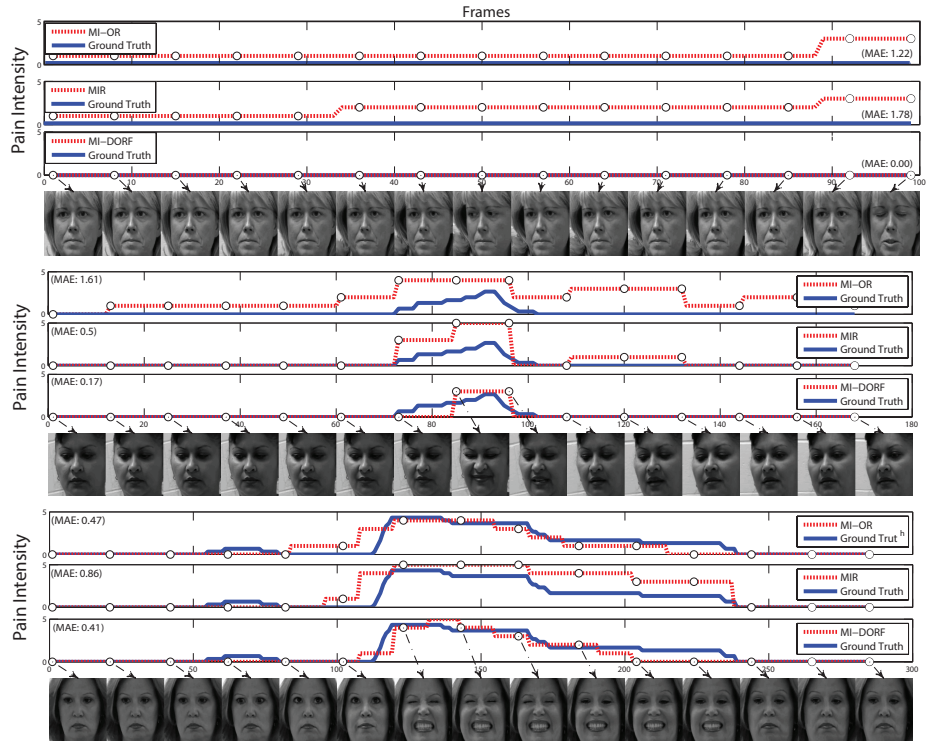}
  \end{center}
  \caption{Visualization of the pain intensity predictions at frame-level for MI-OR, MIR and the proposed MI-DORF method. From top to bottom, three sequences with ground-truth where MI-DORF  predicted the sequence labels: 0, 3 \& 5 respectively.}
  \label{fig:unbc_results_qualitative}
\end{figure}

Finally, in Fig. \ref{fig:unbc_results_qualitative}, we show more qualitative results comparing predictions of MI-OR, MIR and MI-DORF. The shown example sequences depict image frames along with the per-frame annotations and those obtained by compared models, using the adopted weakly-supervised setting (thus, only bag labels are provided). First, we note that all methods succeed to capture the segments in the sequences where the intensity changes occur, as given by the frame-level ground truth. However, note that MI-DORF achieves more accurate localization of the pain activations and prediction of their actual intensity. This is also reflected in terms of the MAE depicted, showing clearly that the proposed outperforms the competing methods on target sequences.

\section{Conclusions}
In this work, we introduced  MI-DORF for the task of Multi-Instance-Ordinal Regression. This is the first MI approach that imposes an ordinal structure on the bag labels, and also attains dynamic modeling of temporal sequences of corresponding ordinal instances. In order to perform inference in the proposed model, we have developed an efficient algorithm with similar computational complexity to that of the standard forward-backward method - despite the fact that we model high-order potentials modelling the MIOR assumption. We demonstrated on the task of weakly supervised pain intensity estimation that the proposed model can successfully unravel the (ordinal) instance labels by using only the (ordinal) bag labels. We also showed that this approach largely outperforms related MI approaches -- all of which fail to efficiently account for either temporal or ordinal, or both types of structure in the target data. 
 
\vspace{3mm}
\noindent {\bf Acknowledgement}.
This paper is part of a project that has received funding from the European Union's Horizon 2020 research and innovation programme under grants agreement no. 645012 ( KRISTINA ), no. 645094 ( SEWA ) and no. 688835 ( DE-ENIGMA ). Adria Ruiz would also like to acknowledge Spanish Government to provide support under grant FPU13/01740.

\bibliographystyle{splncs}

\input{output.bbl}
\end{document}

%% file: baselineresults.tex
\begin{table}[t]
\centering
\begin{tabular}{ccccccccc}
 & \multicolumn{3}{c}{\textbf{Frame-Level}} & \multicolumn{5}{c}{\textbf{Sequence-level}} \\ \cline{2-9} 
 & \textbf{CORR} & \textbf{MAE} & \multicolumn{1}{c|}{\textbf{ICC}} & \textbf{CORR} & \textbf{MAE} & \textbf{ICC} & \textbf{ACC} & \textbf{F1} \\ \hline
\rowcolor[HTML]{EFEFEF} 
\textbf{SIL-OR} & 0.77 & 1.40 & \multicolumn{1}{c|}{\cellcolor[HTML]{EFEFEF}0.71} & 0.85 & 1.43 & 0.57 & 0.26 & 0.19 \\
\rowcolor[HTML]{FFFFFF} 
\textbf{MI-OR} & 0.82 & 0.54 & \multicolumn{1}{c|}{\cellcolor[HTML]{FFFFFF}0.80} & 0.92 & 0.58 & 0.91 & 0.48 & 0.44 \\
\rowcolor[HTML]{EFEFEF} 
\textbf{HCORF \cite{kim2010hidden}} & 0.81 & 1.33 & \multicolumn{1}{c|}{\cellcolor[HTML]{EFEFEF}0.80} & 0.94 & 0.28 & 0.94 & 0.74 & 0.74 \\
\rowcolor[HTML]{FFFFFF} 
{\color[HTML]{000000} \textbf{HCRF \cite{quattoni2007hidden}}} & {\color[HTML]{000000} 0.49} & {\color[HTML]{000000} 1.41} & \multicolumn{1}{c|}{\cellcolor[HTML]{FFFFFF}{\color[HTML]{000000} 0.42}} & {\color[HTML]{000000} 0.93} & {\color[HTML]{000000} 0.36} & {\color[HTML]{000000} 0.92} & {\color[HTML]{000000} 0.67} & {\color[HTML]{000000} 0.66} \\
\rowcolor[HTML]{EFEFEF} 
{\color[HTML]{000000} \textbf{MIR  \cite{hsu2014augmented}}} & {\color[HTML]{000000} 0.79} & {\color[HTML]{000000} 0.58} & \multicolumn{1}{c|}{\cellcolor[HTML]{EFEFEF}{\color[HTML]{000000} 0.78}} & {\color[HTML]{000000} 0.92} & {\color[HTML]{000000} 0.42} & {\color[HTML]{000000} 0.91} & {\color[HTML]{000000} 0.61} & {\color[HTML]{000000} 0.61} \\
\rowcolor[HTML]{FFFFFF} 
{\color[HTML]{000000} \textbf{MI-HCRF}} & {\color[HTML]{000000} 0.77} & {\color[HTML]{000000} 0.75} & \multicolumn{1}{c|}{\cellcolor[HTML]{FFFFFF}{\color[HTML]{000000} 0.67}} & {\color[HTML]{000000} 0.93} & {\color[HTML]{000000} 0.43} & {\color[HTML]{000000} 0.93} & {\color[HTML]{000000} 0.59} & {\color[HTML]{000000} 0.58} \\
\rowcolor[HTML]{EFEFEF} 
{\color[HTML]{000000} \textbf{MI-DORF}} & {\color[HTML]{000000} \textbf{0.86}} & {\color[HTML]{000000} \textbf{0.39}} & \multicolumn{1}{c|}{\cellcolor[HTML]{EFEFEF}{\color[HTML]{000000} \textbf{0.85}}} & {\color[HTML]{000000} \textbf{0.96}} & {\color[HTML]{000000} \textbf{0.20}} & {\color[HTML]{000000} \textbf{0.96}} & {\color[HTML]{000000} \textbf{0.80}} & {\color[HTML]{000000} \textbf{0.80}} \\ \hline
\end{tabular}
\caption{The performance of different methods obtained on the synthetic data.}
\label{tab:synth_results}
\vspace{-0.5cm}
\end{table}

%% file: unbcresults.tex
\begin{table}[t]
\centering
\begin{tabular}{ccccccccc}
\cline{2-9}
 & \multicolumn{3}{c}{{\color[HTML]{000000} \textbf{Frame-Level}}} & \multicolumn{5}{c}{{\color[HTML]{000000} \textbf{Sequence-level}}} \\ \cline{2-9} 
 & {\color[HTML]{000000} \textbf{CORR}} & {\color[HTML]{000000} \textbf{MAE}} & \multicolumn{1}{c|}{{\color[HTML]{000000} \textbf{ICC}}} & {\color[HTML]{000000} \textbf{CORR}} & {\color[HTML]{000000} \textbf{MAE}} & {\color[HTML]{000000} \textbf{ICC}} & {\color[HTML]{000000} \textbf{ACC}} & {\color[HTML]{000000} \textbf{F1}} \\ \hline
\rowcolor[HTML]{EFEFEF} 
{\color[HTML]{000000} \textbf{SIL-OR}} & 0.31 & 1.67 & \multicolumn{1}{c|}{\cellcolor[HTML]{EFEFEF}0.22} & 0.59 & 1.52 & 0.56 & 0.19 & 0.16 \\
\rowcolor[HTML]{FFFFFF} 
{\color[HTML]{000000} \textbf{MI-OR}} & 0.39 & 0.76 & \multicolumn{1}{c|}{\cellcolor[HTML]{FFFFFF}0.28} & 0.64 & 1.01 & 0.63 & 0.39 & 0.31 \\
\rowcolor[HTML]{EFEFEF} 
{\color[HTML]{000000} \textbf{HCORF \cite{kim2010hidden}} } & 0.24 & 1.92 & \multicolumn{1}{c|}{\cellcolor[HTML]{EFEFEF}0.12} & 0.30 & 1.36 & 0.30 & 0.39 & 0.19 \\
\rowcolor[HTML]{FFFFFF} 
{\color[HTML]{000000} \textbf{HCRF \cite{quattoni2007hidden}}  } & {\color[HTML]{000000} 0.09} & {\color[HTML]{000000} 2.29} & \multicolumn{1}{c|}{\cellcolor[HTML]{FFFFFF}{\color[HTML]{000000} 0.05}} & {\color[HTML]{000000} 0.26} & {\color[HTML]{000000} 1.52} & {\color[HTML]{000000} 0.26} & {\color[HTML]{000000} 0.29} & {\color[HTML]{000000} 0.13} \\
\rowcolor[HTML]{EFEFEF} 
{\color[HTML]{000000} \textbf{MIR  \cite{hsu2014augmented}}} & {\color[HTML]{000000} 0.35} & {\color[HTML]{000000} 0.84} & \multicolumn{1}{c|}{\cellcolor[HTML]{EFEFEF}{\color[HTML]{000000} 0.24}} & {\color[HTML]{000000} 0.63} & {\color[HTML]{000000} 0.94} & {\color[HTML]{000000} 0.63} & {\color[HTML]{000000} 0.41} & {\color[HTML]{000000} 0.30} \\
\rowcolor[HTML]{FFFFFF} 
{\color[HTML]{000000} \textbf{MILBoost \cite{sikka2013weakly}}  \ } & {\color[HTML]{000000} 0.28} & {\color[HTML]{000000} 1.77} & \multicolumn{1}{c|}{\cellcolor[HTML]{FFFFFF}{\color[HTML]{000000} 0.11}} & {\color[HTML]{000000} 0.38} & {\color[HTML]{000000} 1.7} & {\color[HTML]{000000} 0.38} & {\color[HTML]{000000} 0.3} & {\color[HTML]{000000} 0.2} \\
\rowcolor[HTML]{EFEFEF} 
{\color[HTML]{000000} \textbf{MI-HCRF}} & {\color[HTML]{000000} 0.17} & {\color[HTML]{000000} 1.45} & \multicolumn{1}{c|}{\cellcolor[HTML]{EFEFEF}{\color[HTML]{000000} 0.11}} & {\color[HTML]{000000} 0.26} & {\color[HTML]{000000} 1.69} & {\color[HTML]{000000} 0.26} & {\color[HTML]{000000} 0.28} & {\color[HTML]{000000} 0.21} \\
\rowcolor[HTML]{FFFFFF} 
{\color[HTML]{000000} \textbf{MI-DORF}} & {\color[HTML]{000000} \textbf{0.40}} & {\color[HTML]{000000} \textbf{0.19}} & \multicolumn{1}{c|}{\cellcolor[HTML]{FFFFFF}{\color[HTML]{000000} \textbf{0.40}}} & {\color[HTML]{000000} \textbf{0.67}} & {\color[HTML]{000000} \textbf{0.80}} & {\color[HTML]{000000} \textbf{0.66}} & {\color[HTML]{000000} \textbf{0.52}} & {\color[HTML]{000000} \textbf{0.34}} \\ \hline
\end{tabular}
\caption{The performance of different methods obtained on the UNBC Database.}
\label{tab:unbc_results}
\vspace{-0.5cm}
\end{table}